\documentclass[preprints,article,accept,moreauthors,pdftex]{Definitions/mdpi} 
\usepackage[ruled]{algorithm2e}
\usepackage[para,online,flushleft]{threeparttable}
\graphicspath{{images/},{.}}
\DeclareGraphicsExtensions{.pdf,.jpeg,.png,.jpg}
\usepackage{fancyhdr}
\firstpage{1} 
\pubvolume{xx}
\issuenum{1}
\articlenumber{5}
\pubyear{2020}
\copyrightyear{2020}
\history{Received: date; Accepted: date; Published: date}

\pdfoutput=1

\Title{Exploration of Optimized Semantic Segmentation Architectures for edge-Deployment on Drones}

\Author{Vivek Parmar $^{1,\dagger}$, Narayani Bhatia $^{1,\dagger}$, Shubham Negi $^{1}$ and Manan Suri $^{1,2}$*}

\AuthorNames{Vivek Parmar, Narayani Bhatia, Shubham Negi and Manan Suri}

\address{%
$^{1}$ \quad Department of Electrical Engineering, Indian Institute of Technology Delhi; manansuri@iitd.ac.in\\
$^{2}$ \quad CYRAN AI Solutions, contact@cyran.in}

\corres{Correspondence: manansuri@iitd.ac.in}

\firstnote{These authors contributed equally to this work.}

\abstract{In this paper, we present an analysis on the impact of network parameters for semantic segmentation architectures in context of UAV data processing. We present the analysis on the DroneDeploy Segmentation benchmark. Based on the comparative analysis we identify the optimal network architecture to be FPN-EfficientNetB3 with pretrained encoder backbones based on Imagenet Dataset. The network achieves IoU score of 0.65 and F1-score of 0.71 over the validation dataset. We also compare the various architectures in terms of their memory footprint and inference latency with further exploration of the impact of TensorRT based optimizations. We achieve memory savings of ~ 4.1x and latency improvement of 10\% compared to Model: FPN and Backbone: InceptionResnetV2.}


\begin{document}


\fancypagestyle{plain}{%
  \renewcommand{\headrulewidth}{0pt}%
  \fancyhf{}%
  \fancyhead[C]{\textcolor{blue}{This manuscript is submitted for review.}}%
  \fancyfoot[C]{\textcolor{blue}{This manuscript is submitted for review.}}%
}
\section{Introduction}
Remote sensing applications have gained immense traction in recent years owing to the advent of high quality acquisition systems, sophisticated processing algorithms and increasingly accurate detection and classification methods enabled by deep learning. Owing to it's rich features, some of the most popular remote sensing tasks rely on semantic segmentation to assign class-wise labels to each pixel in the frame. Semantic segmentation algorithms are used for a plethora of applications such as anomaly detection, event detection, land use cover change, etc \cite{yao2019unmanned}. Unmanned Aerial Vehicles (UAVs) have enabled capture of ultra-high resolution data \cite{yao2019unmanned} due to characteristics like low-cost, flexibility and low-flying altitude thus leading to increasing interest in the field. 

In the context of UAV-centric deep learning applications, there has been some work around autonomous navigation \cite{fraga2019review}, object tracking \cite{aguilar2017pedestrian}, change detection \cite{avola2018uav}, semantic segmentation \cite{Yang_2020}. In the context of segmentation, some papers use standalone networks such as FCN, U-Net\cite{Girisha_2019}, SegNet \cite{Yang_2020, lobo2020applying}, DeepLabV3+\cite{Girisha_2019,lobo2020applying}, Conditional Random Fields \cite{Girisha_2019}, Markov Random Fields \cite{li2017semantic}, Adverserial Networks\cite{li2019road}, while some have experimented with Hybrid Networks such as FCN-AlexNet\cite{Yang_2020}, FCN-ConvLSTM \cite{Wang_2019}, DenseCNN with RNN \cite{rahnemoonfar2018flooded} or Ensemble Networks such as ensemble of ConvNets \cite{nogueira2017semantic}. Some papers have used additional information such as Digital Surface Models \cite{zhang2019urban} or applied post-processing on segmentation results with overlay techniques and probabilistic graphical models\cite{dlreview}.

In recent times, deep learning implementations have been employed for real-time recognition and tracking on drones \cite{neurala} with cloud-based processing. With resource-hungry Deep Convolutional Neural Networks (DCNNs) breaking accuracy ceilings, a significant aspect of the viability relies on reduction of costs in terms of network communication and computational energy. Hence UAV systems warrant use of new-age edge AI (artificial intelligence) hardware accelerators such as edge-GPU (Graphics Processing Units) \cite{Cass_2020}, edge-TPU (Tensor Processing Units) \cite{edgetpu}, specialized ASIC (Application specific Integrated Circuits) \cite{Puglia_2016, parmar_hsi_ijcnn}, which help achieve faster inference time, lightweight deployment and low-power localized processing. Studies proposing embedded deployment in the remote sensing domain have been limited to object detection \cite{barba2020deep,kyrkou2018dronet}, scene classification \cite{kyrkou2019deep}, or semantic segmentation for satellite deployment \cite{inria}. 

Despite standalone work in each of these domains, another untapped aspect for realizing the potential of UAV applications is real time edge AI deployment of segmentation algorithms on UAV platforms. Such applications warrant solutions optimizing all parameters viz. accuracy, latency, and memory thus needing extensive design exploration which is the primary motivation behind this study.

To this purpose, we present a first-of-its-kind intensive algorithmic-hardware exploration for performing segmentation using UAV images based multi-class dataset in a resource-efficient manner for future-ready edge-AI deployments. Key contributions of the paper are listed below:
\begin{enumerate}
    \item Detailed performance benchmarking of standard segmentation models on a new multi-class UAV segmentation dataset (DroneDeploy \cite{dronedeploy}).
    \item First demonstration of EfficientNet based Semantic Segmentation in context of UAV images.
    \item First hardware-software co-optimization study for semantic segmentation in context of UAV images.
\end{enumerate}

\section{Materials and Methods}

\subsection{UAV Datasets}\label{dataset}
\begin{figure}[ht]
\centering
\includegraphics[width=0.9\linewidth]{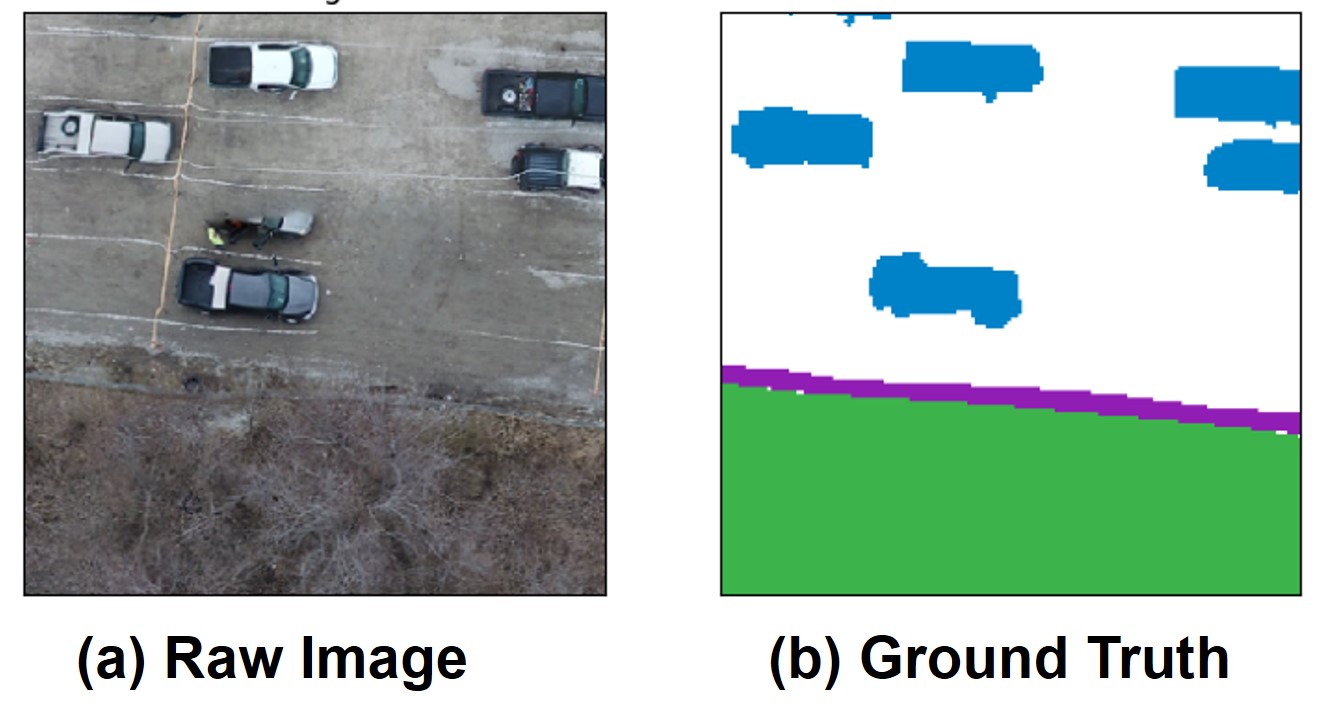}
\caption{Example of image chips used for training from the DroneDeploy dataset\cite{dronedeploy}. (a) RGB image (b) Ground Truth label map.}
\label{pred}
\end{figure}

While there exist many UAV video datasets \cite{Girisha_2019, avola2018uav, uavid_isprs}, UAV static images datasets are typically more application oriented \cite{Yang_2020}, and hence suited for object detection applications \cite{stanforddronedataset}. Further they also generally have lower number of annotated classes \cite{inria}. 
For the purpose of this study, we have used DroneDeploy Dataset\cite{dronedeploy}, comprising of 55 RGB images, along with single-channel elevation maps and label maps. The label maps are annotated with 7 classes - namely Building, Clutter, Vegetation, Water, Ground, Car and `Ignore' - the last class referring to missing pixels/ boundaries. The ground resolution is 10 cm/pixel. For this study, we have only used raw RGB TIFFs, in order to demonstrate generalized capability without the need of additional channels such as elevation (as in the case of this dataset) or hyper-spectral bands (as in the case of other UAV datasets) due to relatively high costs of lightweight multispectral cameras \cite{yao2019unmanned}. A description of the class-wise distribution alongwith color map is provided in Table \ref{table_class_desc}. Maximum image size available in the dataset is 637 MB. Fig. \ref{pred} shows an image chip and corresponding label map from the dataset.

\begin{table}[ht]
\caption{Distribution of class samples in the dataset}
\label{table_class_desc}
\centering
\begin{tabular}{ccc}
\toprule
Class & Color Code & Percentage of Pixels\\
\midrule
Building & Red & 5.6\% \\
\midrule
Clutter & Purple & 2\%\\
\midrule
Vegetation & Green & 10.43\% \\
\midrule
Water & Orange & 1.2\%\\
\midrule
Ground & White & 37.7\%\\
\midrule
Car & Blue & 0.38\%\\
\midrule
Ignore &  Magenta & 42.7\% \\
\bottomrule
\end{tabular}
\end{table}

For training, we divided images in the dataset into non-overlapping chips with chip sizes of 300$\times$300 for UNet, FPN, LinkNet, and 384$\times$384 for PSPNet with ensured presence of at least 1 class of interest. Resulting training dataset consists of 5887 and 3893 chips of sizes 300$\times$300 and 384$\times$384 respectively. Basic augmentations such as Horizontal and Vertical Flips as well as Rotation operations in the range of 180$^\cdot$ were applied during training in order to artificially increase size of the dataset as well as provide better generalization. The ratio of training and validation data is chosen as 85\%:15\%.

\subsection{Semantic Segmentation Architectures}\label{arch} 

In this work, we experiment with a variety of models, backbones, hyper-parameters and training variations before arriving at an accurate model and further try to optimize the architecture for embedded implementation. Segmentation models and encoder-backbones investigated for this study are listed in Table \ref{table_model_bb}. 
Structurally, all semantic segmentation architectures are similar and comprise of an encoder-decoder network where the encoder network performs feature extraction and decoder performs localization of spatial features \cite{inria}. However, it is the method of combining the two sets of information - spatial and feature, which demarcates them. Four models were taken into consideration for this study: (i) UNet \cite{unet}, (ii) LinkNet \cite{linknet}, (iii) Pyramid Scene Parsing Network (PSPNet) \cite{pspnet}, (iv) Feature Pyramid Networks (FPN) \cite{fpn}. 

UNet is one of the most fundamental semantic segmentation networks. It was originally intended to be used on biomedical images, however it finds increasing relevance in nearly all areas of interest today including remote sensing \cite{inria}. In case of UNet, the encoder is used for multi-level feature extraction and the decoder combines learnt features and resolution through a sophisticated stacking, taking both localization and feature representation into account\cite{unet}. LinkNet tweaks the UNet structure by adding the upsampled feature representation with resolution information instead of concatenating. The other two Pyramid Networks attempt to form a Pyramid structure. PSPNet achieves this by creating a pyramid by variably pooling the lowest downsampled map, resulting in a vast collection of spatial resolutions used to enrich the features. On the other hand, FPN works by creating two pyramids, and combines them to generate feature-rich segmentation maps at each level.

Three encoder-backbones have been considered for this study: (i) EfficientNetB3 \cite{tan2019efficientnet}, (ii) InceptionResnetV2 \cite{szegedy2017inception}, (iii) MobileNetV2 \cite{sandler2018mobilenetv2}. The prime motivation for the choice of backbones was to ensure an exhaustive exploration. For analyzing trade-off between memory and accuracy over the complete design space, we have selected three backbones as representative workloads.
InceptionResNetV2, a hybrid of two sophisticated networks, exhibits high accuracy but is computationally heavy, whereas MobileNetV2 despite having lower accuracy enables light-weight edge AI computing \cite{bianco2018benchmark}. EfficientNet architectures use a compound scaling optimization with variable width, depth, resolution in order to optimize Accuracy and FLOPS. We use the EfficientNetB3 architecture in this family which lies approximately in the middle of the spectrum as will be shown in Section \ref{mem_profile}. 

\begin{figure}[H]
\centering
\includegraphics[width=0.9\linewidth]{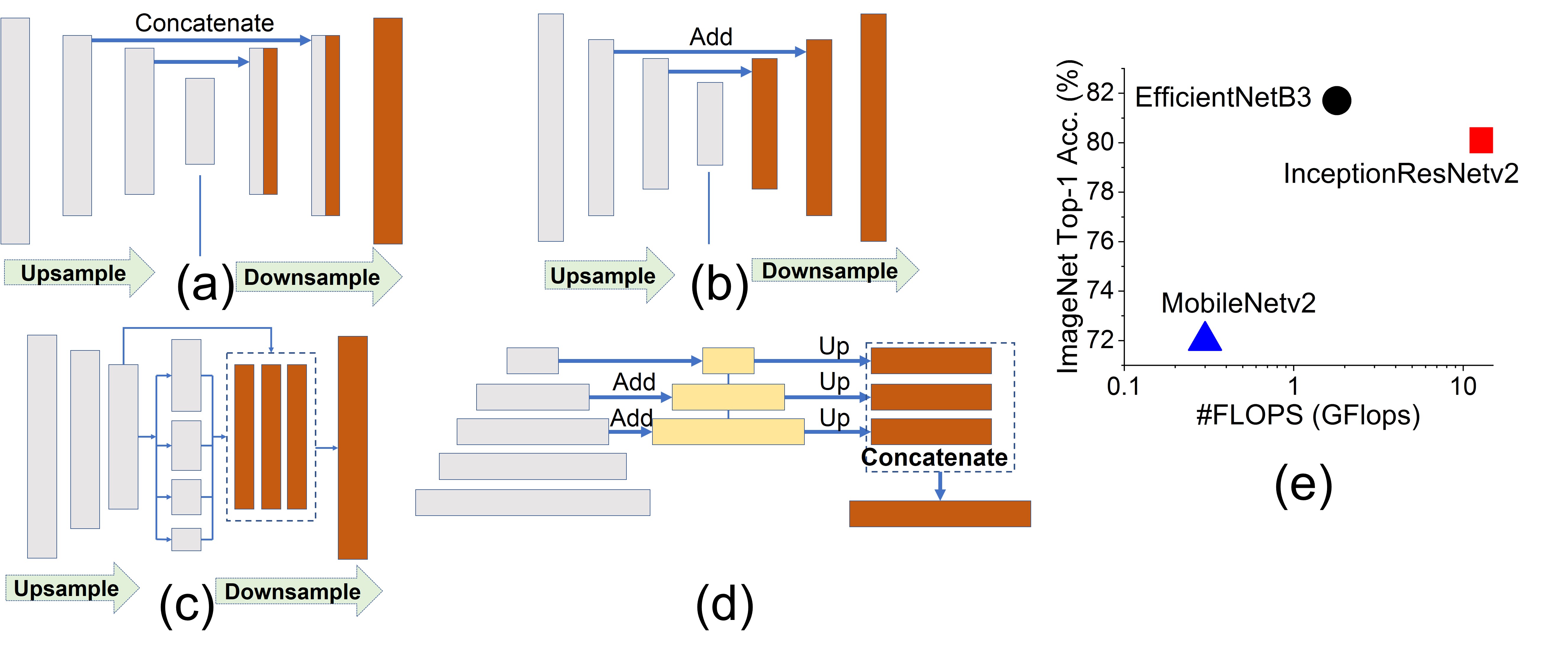}
\caption{Segmentation models investigated in the study: (a) UNet (b) Linknet (c) PSPNet (d) FPN \cite{Yakubovskiy:2019}. (e) Performance benchmark of encoder-backbones investigated in this study (MobileNetv2, EfficientNetB3, InceptionResNetv2) on ImageNet dataset \cite{tan2019efficientnet}.}
\label{seg_models}
\end{figure}

\begin{table}[ht]
\caption{Network architectures used in this study}
\label{table_model_bb}
\centering
\begin{tabular}{cl}
\toprule
Models & UNet, FPN, LinkNet, PSPNet\\
\midrule
Encoder Backbone & MobileNetv2, EfficientNetB3, \\
& Inception ResNetv2\\
\bottomrule
\end{tabular}
\end{table}

\subsection{Experiments}\label{expt}
\begin{figure}[ht]
\centering
\includegraphics[width=0.9\linewidth]{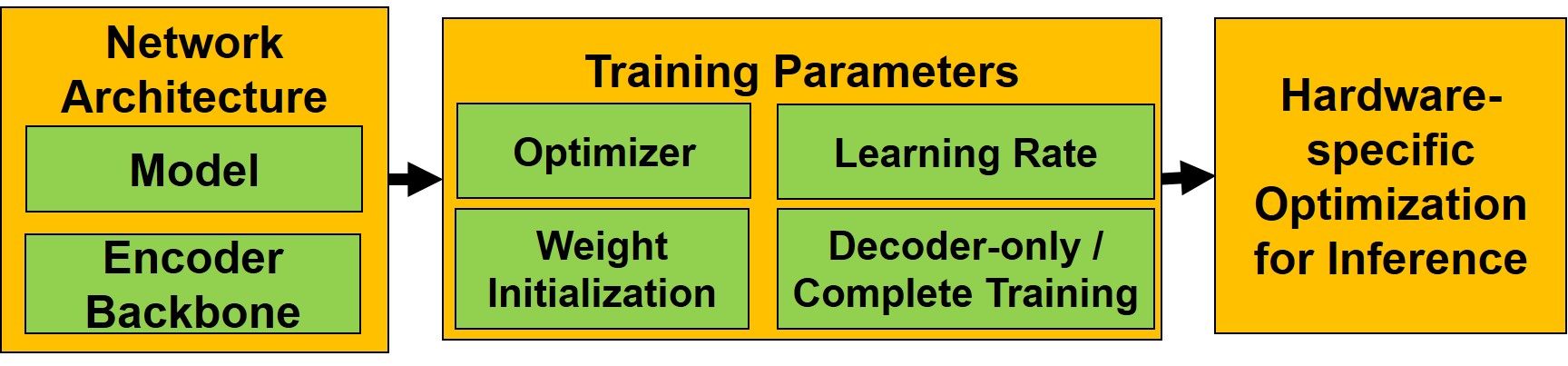}
\caption{Flowchart depicting steps for network parameter exploration.}
\label{flow}
\end{figure}

Fig. \ref{flow} depicts the process of network exploration followed in this study. At the level of network architecture, a total of 12 model-backbone combinations were investigated as shown in Table \ref{table_model_bb}. In case of training parameters, 4 choices viz. optimizer, learning rate policy, weight initialization and decoder-only/ complete training are explored. Further, detailed memory and latency benchmarking is performed for hardware-specific optimization. Experiments performed in the study are based on the Segmentation Models \cite{Yakubovskiy:2019} library based on the Keras framework \cite{chollet2015keras}. For hardware-specfic optimization we utilize NVIDIA TensorRT library\cite{vanholder2016efficient}. A base configuration was selected and for all experiments, where only the parameter of interest was changed. The base configuration is defined in Table \ref{table_cfg}. Two evaluation metrics used for the optimization are defined below.

\begin{enumerate}
    \item IoU Score or the Jaccard Index. 
    \begin{equation}
    IoU (x,y) = \frac{ x \cap y } {x \cup y}
    \end{equation}
    \item F1 Score or the Dice Coefficient ($\beta$ = 1)
    \begin{equation}
    F_1 = 2 \times \frac{ precision * recall } { precision * recall}
    \end{equation}
    where  $precision = \frac{TP}{TP+FP}$,
    $recall = \frac{TP}{TP+FN}$\\
\end{enumerate}

\label{results}

\begin{table}[H]
\caption{Base network architecture used for the Study.}
\label{table_cfg}
\centering
\begin{tabular}{cc}
\toprule
Parameter & Configuration \\
\midrule
Model & UNet \\
\midrule
Backbone & EfficientNetB3 \\
\midrule
Optimizer & AdamW \\
\midrule
Learning Rate Policy & Static Rate (1e-4) \\
\midrule
Weight Initialization & ImageNet \cite{deng2009imagenet} \\
\midrule
Decoder/Complete Training & Complete \\
\midrule
Data Type & FP32 \\
\bottomrule
\end{tabular}
\end{table}

\section{Results}

\subsection{Network architecture exploration}\label{networkexp}

\begin{figure}[H]
\centering
\includegraphics[width=0.92\linewidth]{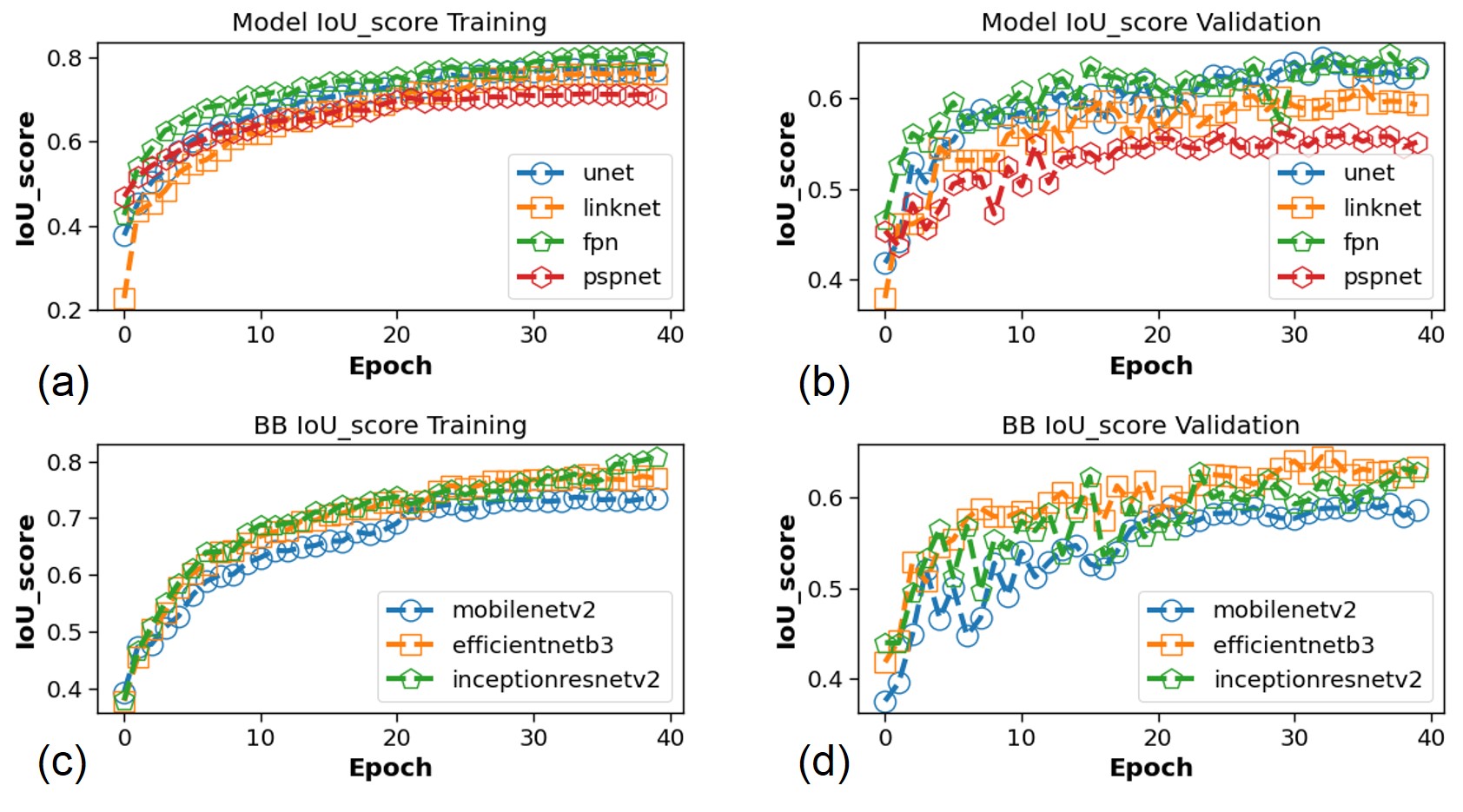}
\caption{Epoch-wise IoU score variation from network architecture experiments for: (i) Model selection with EfficientNetB3 backbone: (a) Training (b) Validation; (ii) Backbone selection with UNet model: (c) Training (d) Validation.}
\label{iou-unet-effnet}
\end{figure}

\begin{figure}[H]
\centering
\includegraphics[width=0.92\linewidth]{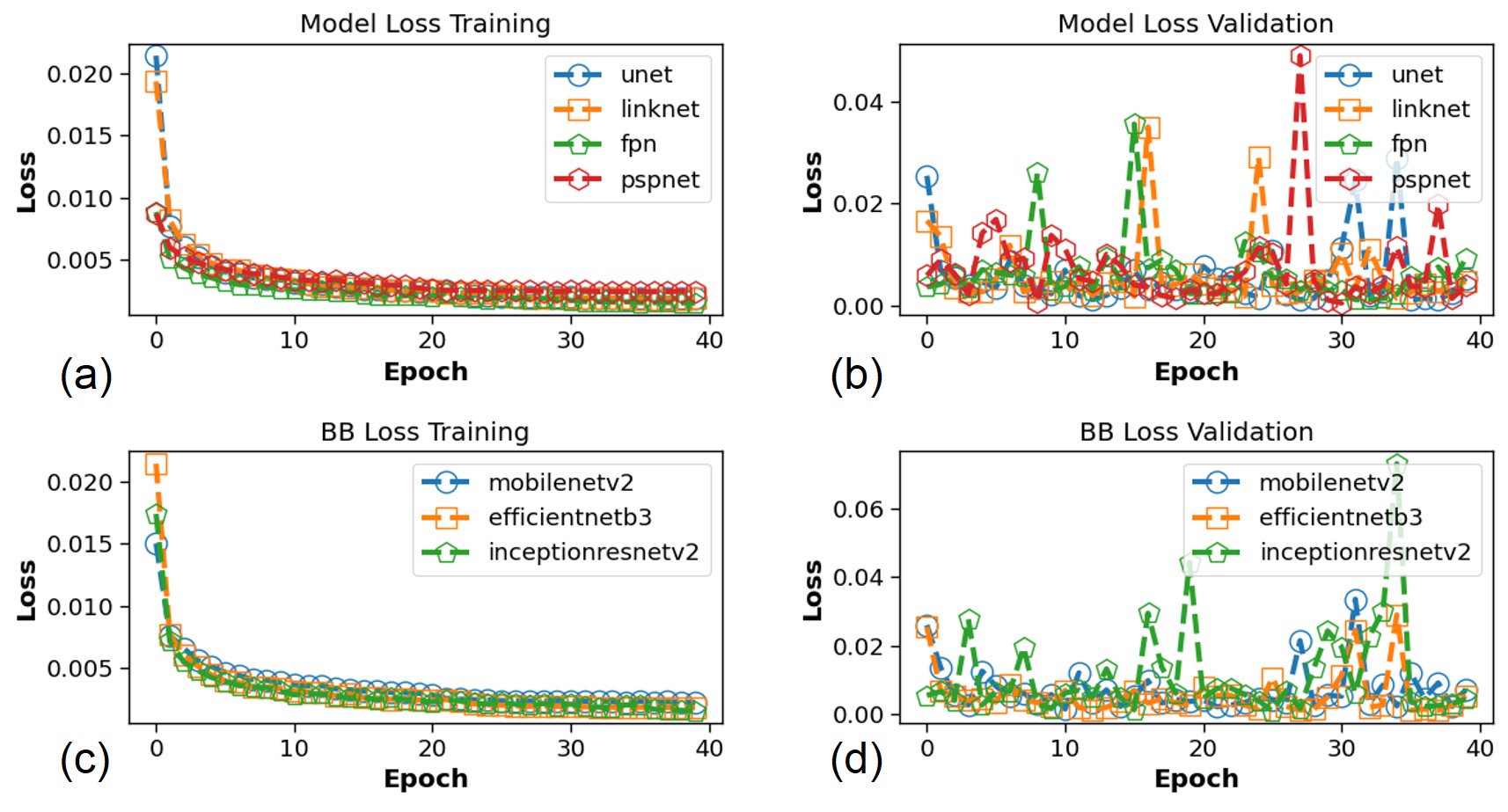}
\caption{Epoch-wise loss variation from network architecture experiments for: (i) Model selection with EfficientNetB3 backbone: (a) Training loss (b) Validation loss; (ii) Backbone selection with UNet model: (c) Training loss (d) Validation loss.}
\label{loss-unet-effnet}
\end{figure}

\begin{figure}[H]
\centering
\includegraphics[width=0.92\linewidth]{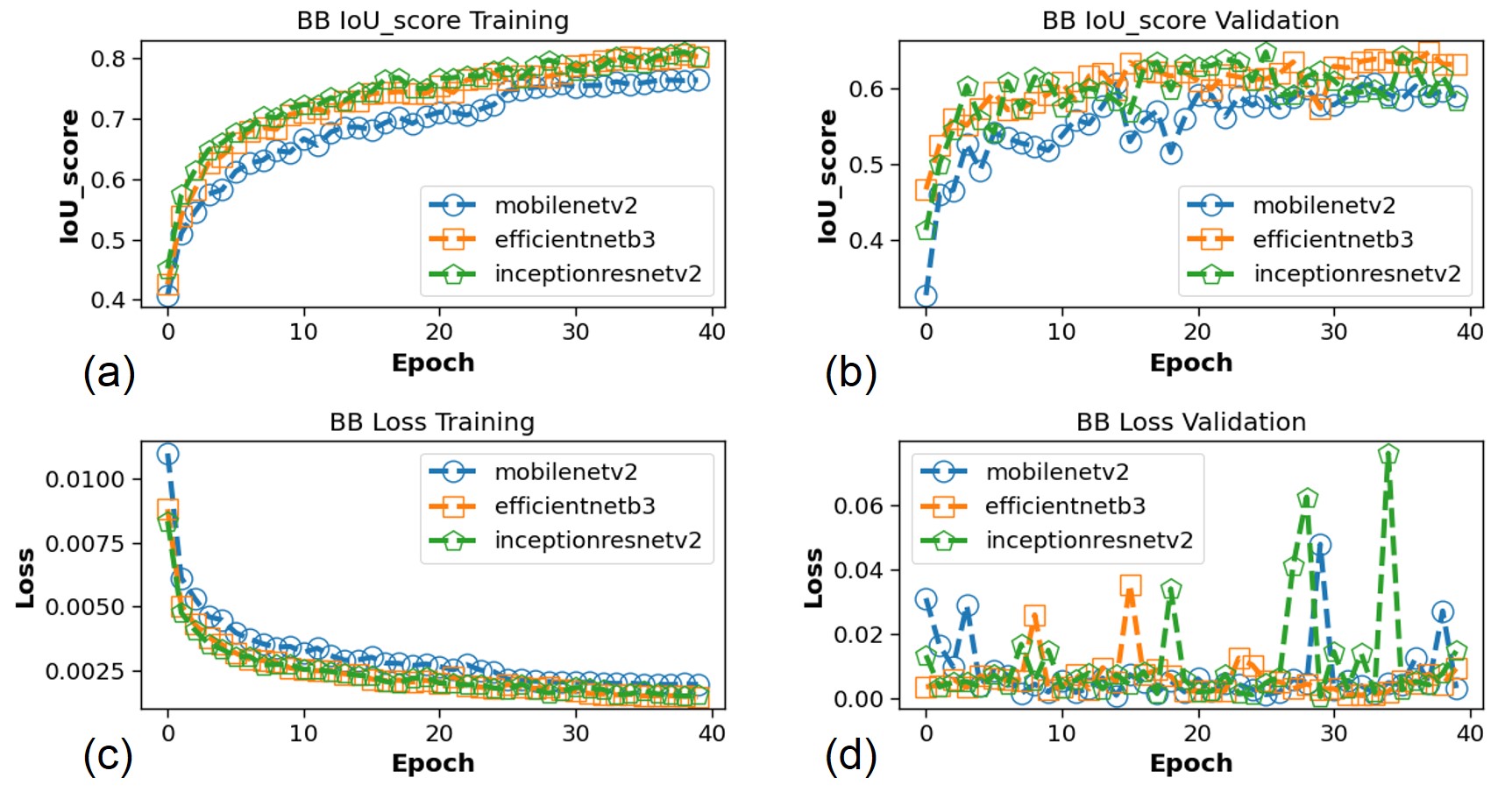}
\caption{Epoch-wise performance analysis of FPN model with backbone variations. IoU score evolution: (a) Training (b) Validation. Loss evolution: (a) Training (b) Validation.}
\label{fpn_allbb}
\end{figure}

Fig. \ref{iou-unet-effnet} and Fig. \ref{loss-unet-effnet} show comparison of performance of different segmentation models with EfficientNetB3 backbone. The best performing Model is found to be FPN. When UNet is selected as the Model, best performing backbone can be taken to be either EfficientNetB3 or InceptionResnetV2 owing to their close performance. Since FPN emerged as a clear choice from the first set of experiments, we then updated the base model to FPN and performed a second set of experiments to compare the performance of encoder-backbones. From Fig. \ref{fpn_allbb} we can observe that EfficientNetB3 shows the best performance and hence is the choice of backbone. This is further validated from results shown in Table \ref{table_iou_modelbb}. FPN with backbone as EfficientNetB3 or InceptionResnetV2 is shown to have highest IoU score, however EfficientNetB3 results in marginally better Validation IoU Score, hence we select it as the backbone. The implications of this choice for hardware will be discussed in detail in Section \ref{mem_profile} and Section \ref{time_opt}.

\begin{table}[H]
\caption{Network architecture exploration results in terms of IoU Scores}
\label{table_iou_modelbb}
\centering
\begin{tabular}{cccc}
\toprule
Model & Backbone & Train IoU & Val IoU \\
\midrule
\multirow{4}{*}{UNet} & EfficientNetB3 & 0.77 & 0.64 \\\cline{2-4}
                   & InceptionResnetV2 & 0.80 & 0.63 \\\cline{2-4}
                   & MobileNetV2 & 0.74 & 0.59 \\
\midrule
\multirow{4}{*}{FPN} & EfficientNetB3 & \textbf{0.81} & \textbf{0.65} \\\cline{2-4}
                   & InceptionResnetV2 & \textbf{0.81} & \textbf{0.65} \\\cline{2-4}
                   & MobileNetV2 & 0.76 & 0.61 \\
\midrule
\multirow{4}{*}{LinkNet} & EfficientNetB3 & 0.76 & 0.61 \\\cline{2-4}
                   & InceptionResnetV2 & 0.78 & 0.63 \\\cline{2-4}
                   & MobileNetV2 & 0.71 & 0.54 \\
\midrule
\multirow{4}{*}{PSPNet} & EfficientNetB3 & 0.71 & 0.56 \\\cline{2-4}
                   & InceptionResnetV2 & 0.74 & 0.58 \\\cline{2-4}
                   & MobileNetV2 & 0.68 & 0.52 \\
\bottomrule
\end{tabular}
\end{table}

\subsection{Training parameter exploration}\label{trainingexp}
Based on the results of the network architecture exploration,
the base configuration is now updated to Model:FPN and Backbone:EfficientNetB3. In these set of experiments, we focus on finding an optimal configuration for the following training parameters:optimizer, learning rate policy, weight initialization, decoder-only/complete training and understanding their impact. Description for each investigated parameter is provided below. 

\subsubsection{Optimizer}\label{optim}
\begin{figure}[H]
\centering
\includegraphics[width=0.92\linewidth]{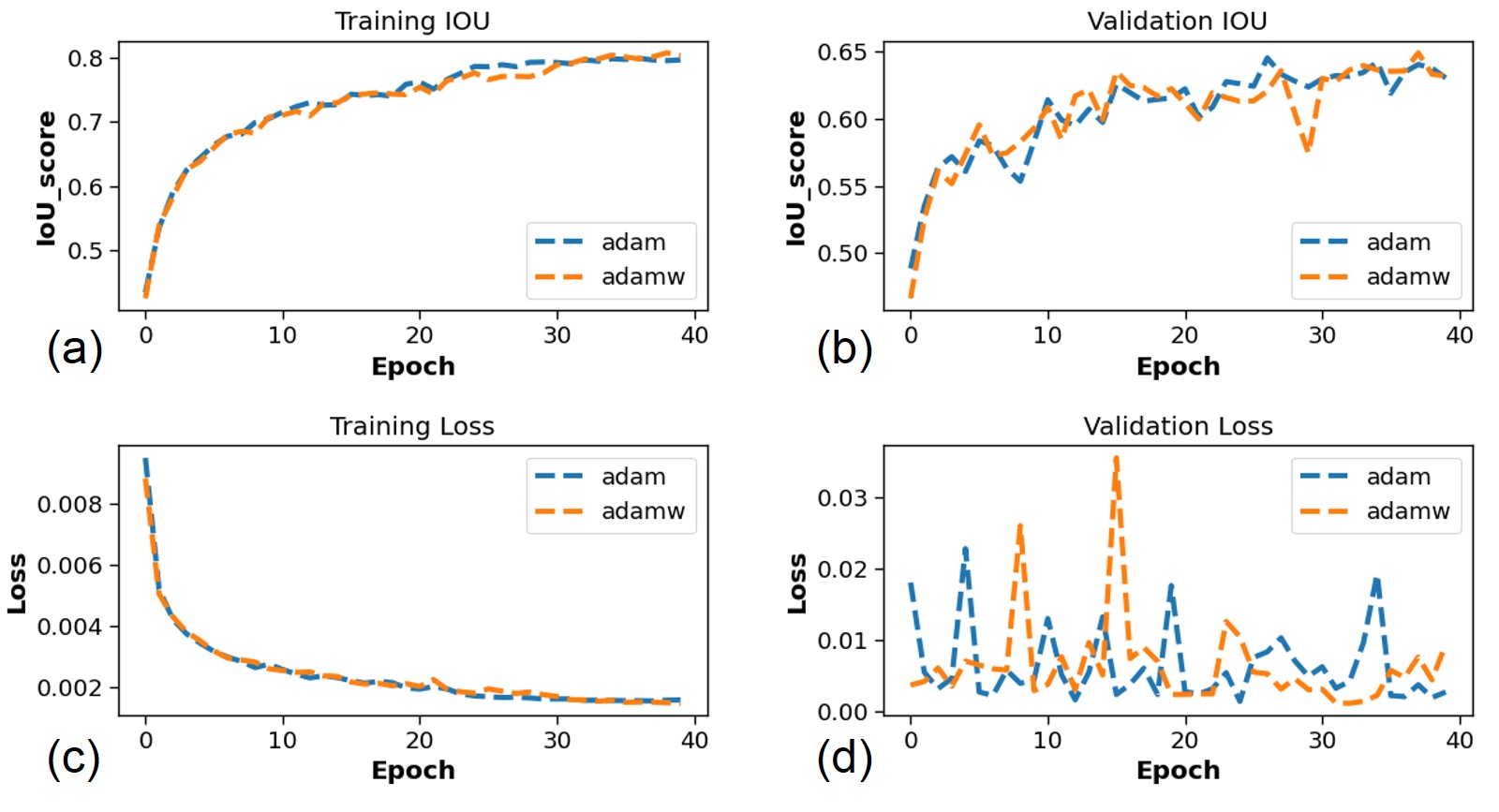}
\caption{Comparison of performance between Adam and AdamW optimizers: (a) Training IoU score (b) Validation IoU score (c) Training loss (d) Validation loss.}
\label{adam_adamw}
\end{figure}

Adam optimizer\cite{kingma2014adam} is one of the most commonly used optimizers used in context of semantic segmentation networks, even finding applications in context of hyperspectral images owing to its high accuracy in classification \cite{paoletti2019deep}. As part of this experiment, we compare the performance of Adam and AdamW \cite{loshchilov2017decoupled} optimizers, wherein AdamW decouples weight decay from the gradient updates. Fig. \ref{adam_adamw} shows better performance for AdamW than Adam, and makes it an optimal choice as the optimizer.

\subsubsection{Learning Rate Policy}\label{lr}
Learning rate (LR) is an important parameter for preventing overfitting or getting stuck in local minima. Hence as part of this experiment we compare performance between two learning rate policies: (i) static LR throughout the training epochs, (ii) LR scheduler based as defined in algorithm \ref{algo1}. 


\begin{algorithm}[H]
\caption{Learning rate scheduler algorithm}
\label{algo1}
\SetAlgoLined
\KwData{Current Epoch $e$,Max Epoch $E_{max}$, Base Learning Rate $LR_{base}$, Decay factor $d$}
\KwResult{New Learning Rate $LR_{new}$}
\eIf{$e<E_{max}/2$}{$LR_{new}=LR_{base}$}
{$LR_{new}=LR_{base}$ $\times$ $e^{(d\times(\frac{E_max}{2}-e)}$}
\end{algorithm}

\begin{figure}[H]
\centering
\includegraphics[width=0.92\linewidth]{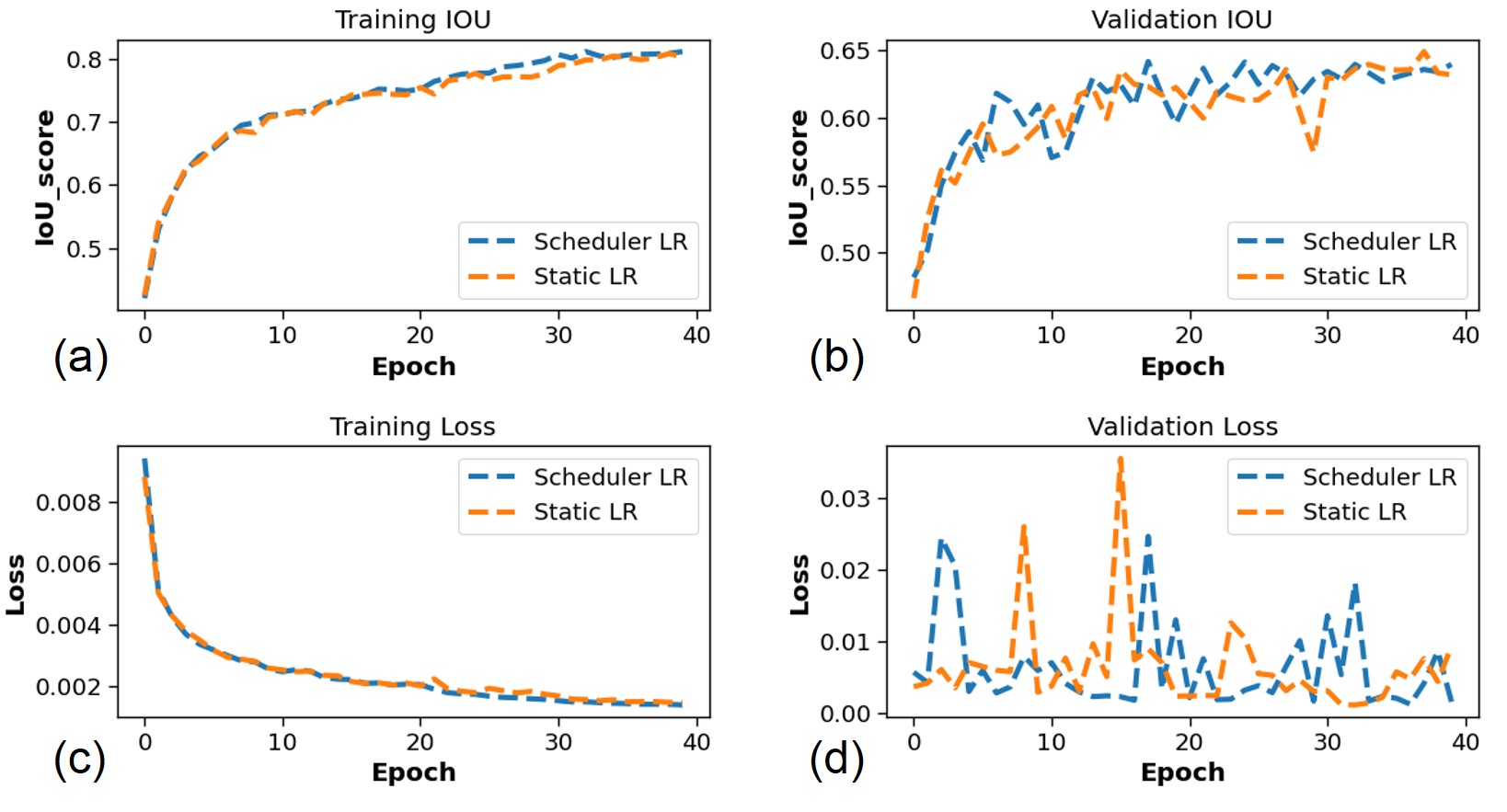}
\caption{Comparison of performance for static and scheduler LR based training: (a) Training IoU score (b) Validation IoU score (c) Training loss (d) Validation loss.}
\label{learningrate}
\end{figure}

Fig. \ref{learningrate} shows that using a LR scheduler policy with an adaptive learning rate performs better compared to a static LR policy. Thus base configuration is updated with a scheduler based learning rate policy. 

\subsubsection{Weight Initialization}\label{wi}
\begin{figure}[H]
\centering
\includegraphics[width=0.92\linewidth]{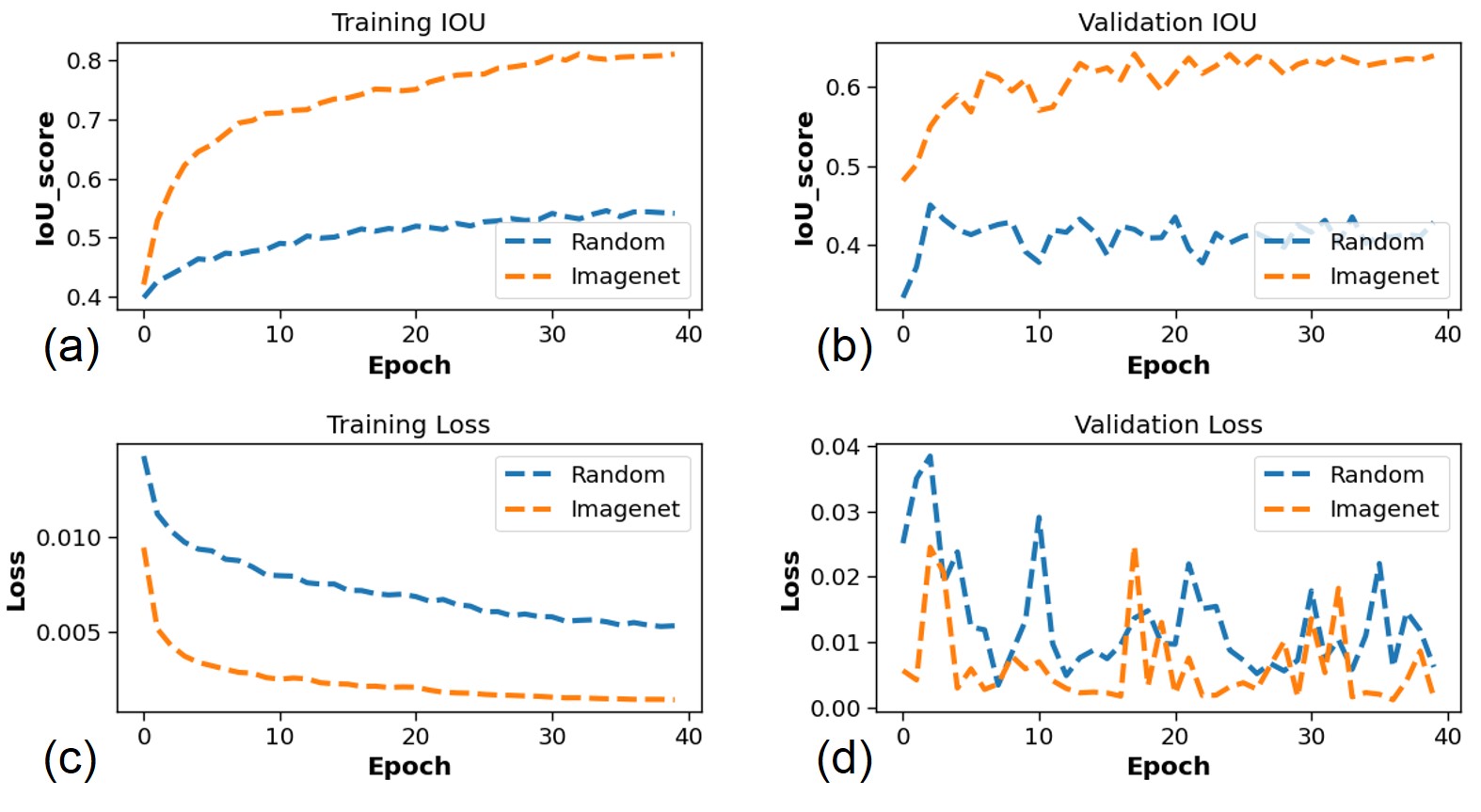}
\caption{Comparison of performance for weight-initialization strategies: (a) Training IoU score (b) Validation IoU score (c) Training loss (d) Validation loss.}
\label{weightinit}
\end{figure}

While using ImageNet weights to initialize pre-training is a widely accepted method to accelerate training for deep learning, we also have to acknowledge the difference in target classes specific to the UAV dataset. Hence we studied the effect of pre-training by ImageNet \cite{deng2009imagenet} in comparison with random weight initialization. Fig. \ref{weightinit} shows that a network pre-trained on ImageNet outperforms random initialization by a huge margin. Hence, we proceed with this weight initialization scheme for the base configuration. 

\subsubsection{Decoder-Only / Complete Training}
\label{dec_or_fulltraining}
\begin{figure}[H]
\centering
\includegraphics[width=0.92\linewidth]{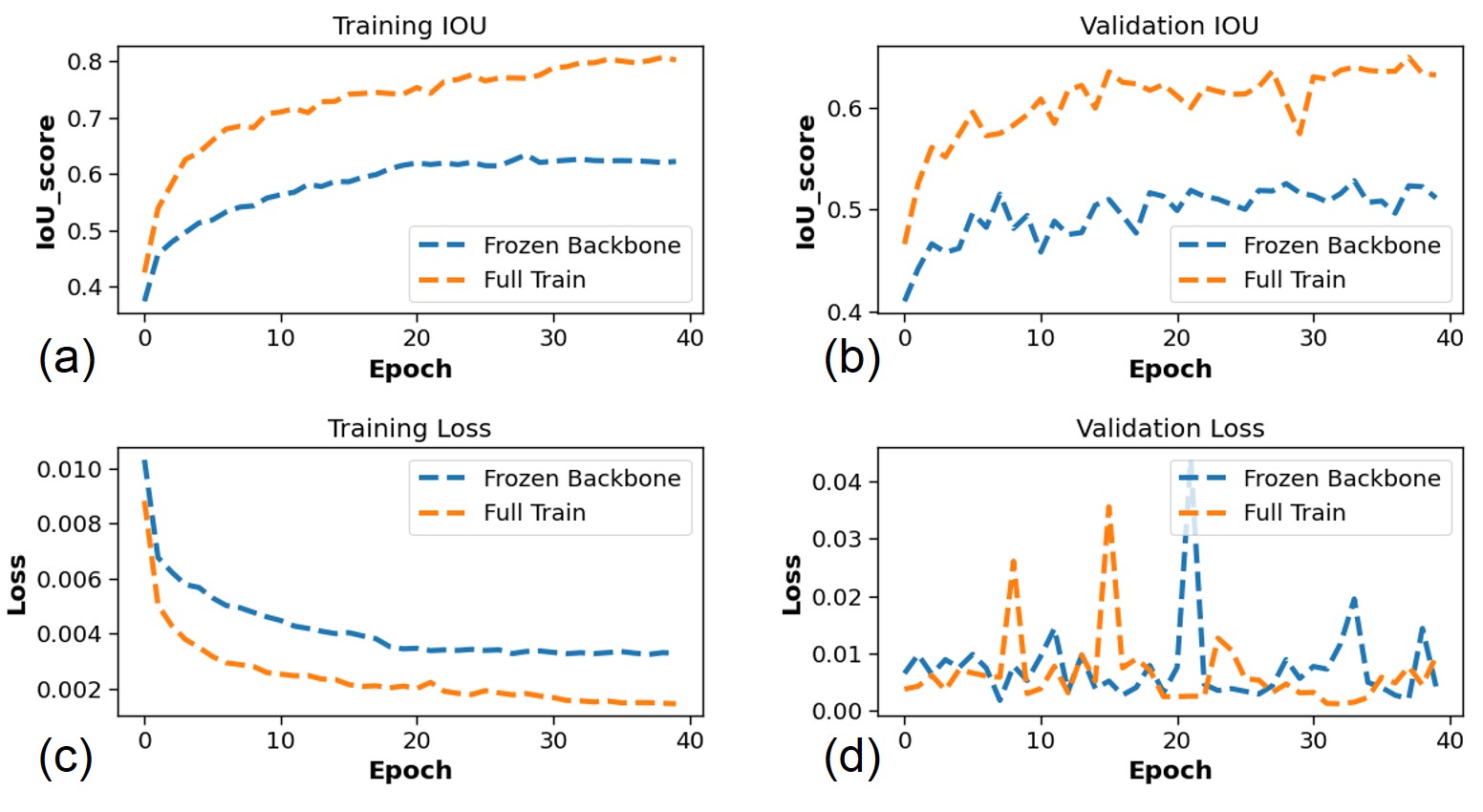}
\caption{Comparison of performance between Complete and Decoder-Only training: (a) Training IoU score (b) Validation IoU score (c) Training loss (d) Validation loss.}
\label{encfreeze}
\end{figure}

As discussed in Section \ref{arch} the semantic segmentation networks have two components: (i) Encoder (ii) Decoder. In the interest of saving the cost of training operations, a commonly used strategy for training these networks is to freeze the encoder weights and only train the decoder \cite{michieli2019incremental}. Another alternative is to train both components (Complete Training). Fig. \ref{encfreeze} shows how complete training outperforms decoder training, possibly owing to the requisite learning for fine-tuned feature extraction from the encoder.

\subsection{Memory Profiling}\label{mem_profile}
So far, we have analyzed methods to update the base configuration in an effort to optimize accuracy. However, real world edge-deployment on any UAV platform will require it to have low memory footprint, low latency and low power. To achieve lightweight implementation in terms of memory footprint, there are two strategies: (i) Network architecture selection to reduce model weights, (ii) Quantization of model weights and computation graph. Both these strategies are discussed further.

\subsubsection{Model Weights}\label{mem_weights}
In this study, we compare the memory consumption of different models' weights as shown in Table \ref{table_mem} in order to select the architecture with the best trade-off between model weight size and validation accuracy.

\begin{table}[H]
\caption{Trade-off between Accuracy and Memory Consumption of various Architectures used in the study}
\label{table_mem}
\centering
\begin{tabular}{cccc}
\toprule
Model & Backbone & Val IOU & Memory (MB) \\
\midrule
\multirow{4}{*}{UNet} & EfficientNetB3 & 0.64 & 68.17 \\\cline{2-4}
                   & InceptionResnetV2 & 0.63 & 236.75 \\\cline{2-4}
                   & MobileNetV2 & 0.59 & 30.7 \\
\midrule
\multirow{4}{*}{FPN} & EfficientNetB3 & \textbf{0.65} & 53.09 \\\cline{2-4}
                   & InceptionResnetV2 & \textbf{0.65} & 219.45  \\\cline{2-4}
                   & MobileNetV2 & 0.61 & 19.90 \\
\midrule
\multirow{4}{*}{LinkNet} & EfficientNetB3 & 0.61 & 52.49 \\\cline{2-4}
                   & InceptionResnetV2 & 0.63 & 220.75 \\\cline{2-4}
                   & MobileNetV2 & 0.54 & 15.81  \\
\midrule
\multirow{4}{*}{PSPNet} & EfficientNetB3 & 0.56 & 7.68 \\\cline{2-4}
                   & InceptionResnetV2 & 0.58 & 13.68 \\\cline{2-4}
                   & MobileNetV2 & 0.52 & \textbf{6.29} \\
\bottomrule
\end{tabular}
\end{table}

\subsubsection{Quantization based Memory Optimization}
\label{mem_dtype}
\begin{figure}[H]
\centering
\includegraphics[width=0.95\linewidth]{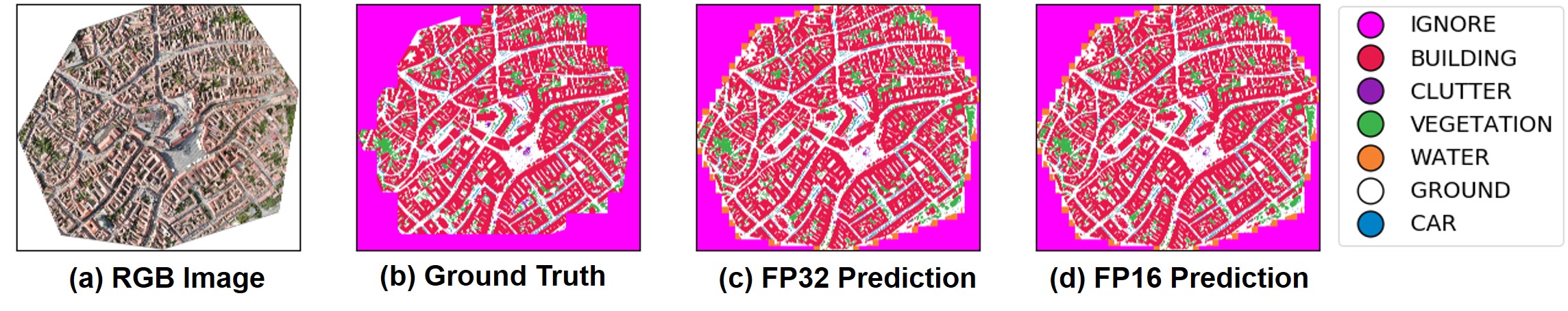}
\caption{Prediction performance of FPN-EfficienNetB3 model: (a) Original RGB image, (b) Ground truth labels, (c) Prediction with FP32 precision (Keras) (d) Prediction with FP16 precision (TF-TRT).}
\label{maximg_pred}
\end{figure}

\begin{figure}[H]
\centering
\includegraphics[width=0.95\linewidth]{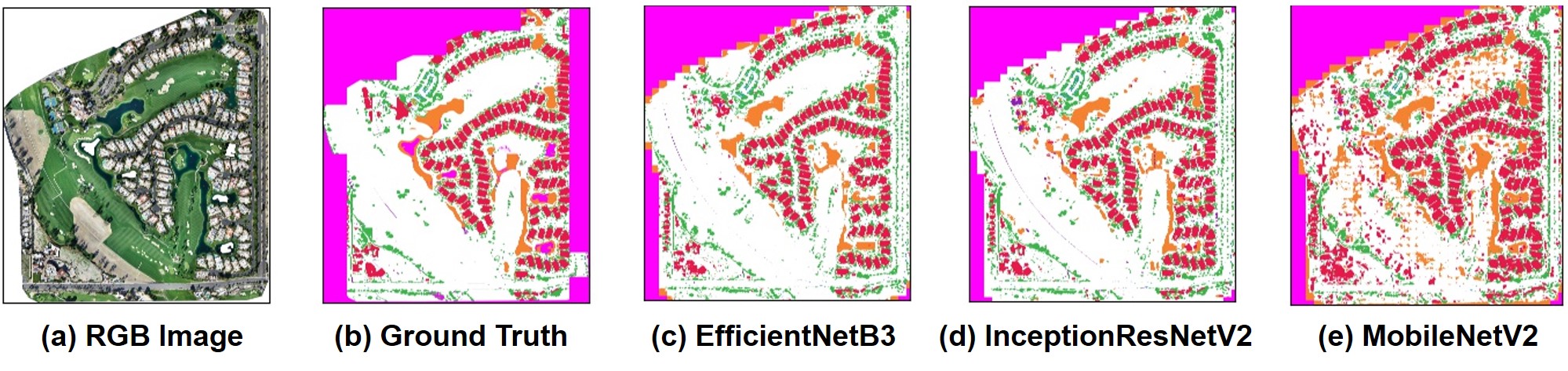}
\caption{Impact of encoder-backbone choice on prediction performance with FPN TF-TRT FP16 model: (a) Original RGB image, (b) Ground truth labels, (c) EfficientNetB3, (d) InceptionResNetv2, (e) MobileNetv2.}
\label{fpn_allbb_pred}
\end{figure}

In order to reduce memory footprint and inference latency, we analyze the impact of quantizing floating-point-32 (FP32) precision network to a more efficient data-type i.e. floating-point-16 (FP16). FP16 quantization is performed based on NVIDIA TensorRT (TRT) \cite{vanholder2016efficient}. Fig. \ref{maximg_pred} demonstrates the prediction performance for the baseline as well as TRT optimized models. We observe that TRT FP16 model performance matches closely with baseline FP32 model, and thus making it a valid candidate for further latency analysis. Fig. \ref{fpn_allbb_pred} shows prediction performance comparison of FP16-architectures with FPN model using all 3 encoder-backbones. As expected, based on earlier analysis (shown in Table \ref{table_iou_modelbb}), MobileNetV2 shows noisy predictions whereas EfficientNetB3 and InceptionResnetV2 are at par with each other. 


\subsection{Latency Profiling}
\label{time_opt}
Table \ref{table_latency} shows the latency analysis for architectures with FPN as model on an input image of size 522 MB, spanning an area of 1.36 $km^{2}$. The analysis is performed on NVIDIA RTX 8000. The updated base model is run on an additional configuration - NVIDIA RTX 2080 Ti since it is more representative of an inference oriented GPU. Keras model is based on FP32 data type, whereas Tensorflow-TensorRT (TF-TRT) implementations are based on FP16 data type. FP16 implementation shows a performance speedup as compared to FP32 thus increasing its viability for deployment. An important point to note here is that while the results presented in Table \ref{table_latency} report latencies for processing a complete ortho-rectified image from the dataset with >1000 slices of 320$\times$320, actual latency for the target platform would be far less depending on sensor resolution. Hence, we present inference latency for image size corresponding to a typical drone sensor (960$\times$480)\cite{dji} in Table \ref{edge-latency}.
Out of the edge-devices referred to in the study, the best performance is obtained for NVIDIA Jetson Nano-Tensorflow framework. For inference time, the throughput metrics of interest are as follows:
\begin{enumerate}
    \item File Size per unit time = 125 kB/second
    \item Area spanned per unit time = 320 $m^{2}$/second
\end{enumerate}

\begin{table}[H]
\caption{Impact of Encoder Backbone on Inference Latency for complete image for FPN architecture}
\label{table_latency}
\centering
\begin{threeparttable}
\begin{tabular}{ccc}
\toprule
Backbone & Keras Model (s) & TF-TensoRT (s) \\
\midrule
EfficientNetB3 & 36.1* & 33.7* \\
\midrule
EfficientNetB3 & 27.8 & 26.3 \\
\midrule
InceptionResnetV2 & 31.5  & 23.3 \\
\midrule
MobileNetV2 & 23.4  & 17.0 \\
\bottomrule
\end{tabular}
\begin{tablenotes}
\footnotesize{*This experiment is performed on NVIDIA RTX 2080 Ti, All other experiments are performed on NVIDIA RTX 8000.}
\end{tablenotes}
\end{threeparttable}
\end{table}

\begin{table}[H]
\caption{Inference latency for typical Drone Image Sensor Resolution (960$\times$480)\cite{dji} using FPN model and EfficientNetB3 Backbone.}
\label{edge-latency}
\centering
\begin{threeparttable}
\begin{tabular}{ccc}
\toprule
    Platform & Framework & Inference Latency (s)\\
    \midrule
    \multirow{2}{*}{Nvidia RTX 2080 Ti} & Tensorflow & 0.07 \\
          & TF-TRT & 0.10 \\
    \midrule
    Xeon CPU & Tensorflow & 0.79\footnotemark \\
    \midrule
    \multirow{2}{*}{Nvidia Jetson Nano} & Tensorflow & 14.40 \\
          & TF-TRT & 19.14\footnotemark \\
    \midrule
    Raspberry Pi 3B+ & Tensorflow & 75.20 \\
\bottomrule
\end{tabular}
\begin{tablenotes}
\footnotesize{$^1$Xeon CPU latency is added from the perspective of overall comparison.}\\
    \footnotesize{$^2$Estimated based on scaling factors derived from \cite{tryolabs}.}
\end{tablenotes}
\end{threeparttable}
\end{table}

\section{Discussion}
Section \ref{networkexp} shows that FPN models with high-level semantic features at every level outperforms UNet models by a considerable margin in terms of validation IoU scores (see Table \ref{table_iou_modelbb}). FPN model with encoder-backbone as EfficientNetB3 has the highest validation IoU score when compared to all other architectures, thus being the preferred choice. The choice of EfficientNetB3 over InceptionResnetV2 stems from the marginally higher IoU score, along with memory footprint considerations as shown in Table \ref{table_mem}. Thus with the choice of the optimal network architecture established, training parameter optimization was explored. In Section \ref{optim} we see that AdamW shows superior performance over Adam optimizer because of incorporation of weight decay (shown in Fig. \ref{adam_adamw}). Section \ref{lr} shows that the scheduler based learning rate policy outperforms the static LR policy by a significant margin due to it's adaptive nature. Section \ref{wi} validates the standard method of weight initialization by ImageNet weights by comparing its performance with regards to random initialization. This is a significant observation in  contrary to the view that remote sensing applications differ considerably from vision applications necessitating departure from standard initializations like ImageNet \cite{deng2009imagenet}. Section \ref{dec_or_fulltraining} shows that complete training allows better feature adaptation than encoder freezing and hence a complete training is followed. Following are some important observations based on trends shown in Section \ref{mem_weights}:

\begin{enumerate}
    \item All models with MobileNetV2 as backbone are lightweight, however, highly underperform when compared to other backbones' Validation IoU.
    \item All InceptionResnetV2 models are very memory intensive, however perform better or at par when compared to other backbones for the same model.
    \item Out of all models, PSPNet has least memory consumption, however, it's performance is also considerably lower than all other models. 
    \item If models with best validation IoU are compared in terms of memory, InceptionResnetV2 is 4x heavier than EfficientNetB3 making it a suboptimal choice for lightweight on-the-edge deployment.
\end{enumerate}

The above trends further explain motivation behind choosing the investigated architectures, since they provide a more complete picture of the tradeoff between accuracy and memory footprint. MobileNetV2 is highly resource efficient for deployment, whereas InceptionResnetV2 performs consistently and accurately, however EfficientNetB3 depicts the tradeoff between accuracy and memory consumption.  

Section \ref{mem_dtype} shows performance of FP 16 optimized models to be at par with FP32 models in terms of accuracy. This observation, compounded with memory footprint savings, motivates the need for investigating the latency performance for the TF-TRT models. 

Section \ref{time_opt} shows lower inference times for TF-TRT optimized models (improvements in the range of $\sim$ 1.05 $\times$ to $\sim$ 1.4 $\times$ except in the case of EfficientNetB3 performed on NVIDIA RTX 2080 Ti, possibly owing to a memory bottleneck issue. NVIDIA Jetson Nano emerges as the preferable choice as based on edge device estimates described in Table \ref{edge-latency}.

\subsection{Future Scope}
While there are several significant observations contributed from the study, following aspects we aim to address in future and on-going work:
\begin{enumerate}
    \item Investigate optimization strategies in order to address UAV dataset imbalance.
    \item TRT edge-AI hardware latency estimates are approximate to a margin of 10\%. Accurate latency and power measurements based on actual deployed characterization will yield more precise values.
    \item Since no gold-standard dataset currently exists for training classifiers for UAV based images as reference, further investigation into impact of weight initialization may be advisable and comparison with ImageNet based initialization.
    \item Power profiling of the networks may give more insight since latency for operations is not representative of energy costs. This would further boost the overall impact of optimization in terms of EDP (Energy Delay Product).
\end{enumerate}

\section{Conclusions}

In this paper, we present a detailed benchmarking study of semantic segmentation models in context of UAV applications on the DroneDeploy dataset. We also present the first demonstration of semantic segmentation based on EfficientNet architectures for remote sensing applications. Based on extensive exploration, the best configuration is found to be: Model: FPN, Backbone: EfficientNetB3, Pretraining: ImageNet weights, Optimizer: AdamW with Learning rate scheduler and complete training which achieves IoU score of 0.65 and F1-score of 0.71 over the validation dataset. We also profile memory usage and latency for each model and optimize them for inference based on TensorRT with FP16 precision. Based on this we achieve memory savings of $\sim$ 4.1$\times$ and latency improvement of $\sim$ 10\% compared to Model: FPN and Backbone: InceptionResnetV2.


\vspace{6pt} 



\authorcontributions{Conceptualization V.P., N.B. and M.S.; Methodology \& Investigation: V.P \& N.B,; SW \& Validation: V.P., N.B., S.N; Writing, Editing and Review: all authors.; Supervision, M.S.}

\funding{This research received no external funding.}

\acknowledgments{Authors would like to acknowledge support provided by CYRAN AI Solutions for this study.}

\conflictsofinterest{The authors declare no conflict of interest.} 

\abbreviations{The following abbreviations are used in this manuscript:\\

\noindent 
\begin{tabular}{@{}ll}
FPN & Feature Pyramid Network \\
PSPNet & Pyramid Scene Parsing Network \\
LR & Learning Rate \\
TRT & Tensor RT
\end{tabular}}




\reftitle{References}

\externalbibliography{yes}
\bibliography{ref}
\end{document}